\documentclass[letterpaper, 10pt, conference]{ieeeconf}  % Comment this line out if you need a4paper

\IEEEoverridecommandlockouts                              % This command is only needed if 
\usepackage{graphicx}
\usepackage{caption}
\usepackage{subcaption}
\usepackage{booktabs}
\usepackage{array}
\usepackage{xcolor}
\usepackage{hyperref}
\hypersetup{
    colorlinks=true,
    linkcolor=red,    % internal links (sections, etc.)
    citecolor=blue,     % citations in the text
    urlcolor=magenta,  % external URLs
}

\usepackage[normalem]{ulem} % for \uline (better-looking underline)
\usepackage{amsmath}
\usepackage{amsfonts}

\newcommand{\prompt}[1]{#1}
\newcommand{\qwentwob}{Qwen3-VL-2B}
\newcommand{\qweneightb}{Qwen3-VL-8B}
\newcommand{\qwensevenb}{Qwen2.5-VL-7B}

\title{\LARGE \bf

Decoding Pedestrian Crossing Intention from Egocentric Vision \\ 
via Vision Language Models
}

\author{Danya Li$^{1}$, Xiang Su$^{2}$, Yan Feng$^{3}$, and Rico Krueger $^{1}$% <-this % stops a space
\thanks{$^{1}$Danya Li and Rico Krueger are with Department of Technology, Management, and Economics, Technical University of Denmark, Denmark
        {\tt\small danli@dtu.dk, rickr@dtu.dk}}%
\thanks{$^{2}$Xiang Su is with Department of Computer Science and Department of Agricultural Sciences, University of Helsinki, Finland
    {\tt\small xiang.su@helsinki.fi}}
\thanks{$^{3}$Yan Feng is with Department of Transport and Planning, Delft University of Technology, Netherlands
    {\tt\small y.feng@tudelft.nl}
}
}

\begin{document}

\maketitle
\thispagestyle{plain}
\pagestyle{plain}

%%%%%%%%%%%%%%%%%%%%%%%%%%%%%%%%%%%%%%%%%%%%%%%%%%%%%%%%%%%%%%%%%%%%%%%%%%%%%%%%
\begin{abstract}

Egocentric vision offers a first-person view of human perception and decision making, yet its potential for traffic-safety prediction remains underexplored. In this work, we study the decoding of pedestrian crossing intentions from short egocentric video clips. We approach this by formulating the task as a closed-ended visual question answering (VQA) problem and leveraging vision language models (VLMs) to predict the pedestrians' intent. We first benchmark three families of state-of-the-art VLMs in a zero-shot setting, finding that they achieve moderate gains over random guessing but exhibit limited higher-level traffic reasoning. Motivated by these findings, we further adapt VLMs to the target task using parameter-efficient fine-tuning. Our results show that the fine-tuned models substantially outperform their zero-shot counterparts and achieve a 9\% accuracy improvement over a specialized transformer-based baseline. Finally, we demonstrate that incorporating additional contextual cues, including ego motion, vehicle motion, and eye gaze, further improves predictive performance. In particular, the fine-tuned \qwentwob\ model guided by eye gaze and ego motion achieves a 14.5\% accuracy improvement over the transformer baseline, establishing a new state of the art for egocentric pedestrian intent decoding.
Code will be made available at \url{https://github.com/danyayay/EgoCross-VLM.git}.

\end{abstract}

\vspace{0.6em}
\noindent
\textbf{Keywords}: pedestrian intention prediction, vision--language models (VLMs), egocentric vision, eye gaze

%%%%%%%%%%%%%%%%%%%%%%%%%%%%%%%%%%%%%%%%%%%%%%%%%%%%%%%%%%%%%%%%%%%%%%%%%%%%%%%%
\section{INTRODUCTION}

% This template provides authors with most of the formatting specifications needed for preparing electronic versions of their papers. All standard paper components have been specified for three reasons: (1) ease of use when formatting individual papers, (2) automatic compliance to electronic requirements that facilitate the concurrent or later production of electronic products, and (3) conformity of style throughout a conference proceedings. Margins, column widths, line spacing, and type styles are built-in; examples of the type styles are provided throughout this document and are identified in italic type, within parentheses, following the example. Some components, such as multi-leveled equations, graphics, and tables are not prescribed, although the various table text styles are provided. The formatter will need to create these components, incorporating the applicable criteria that follow.

Ensuring pedestrian safety in increasingly complex urban environments is critically dependent on accurate anticipation of pedestrian behavior. Traditional approaches have predominantly relied on exocentric perspectives, such as vehicle-mounted or surveillance cameras, which offer a stable and global view of the scene \cite{zhang_pedestrian_2023, landry_predicting_2025}. 
However, these external viewpoints often fail to capture the pedestrian's first-person perception and fine-grained behavioral cues. In contrast, egocentric vision offers direct access to the pedestrian’s line of sight, providing rich information that is crucial to understand intention and action execution \cite{wang_egonav_2024, he_bridging_2026}. 

The growing interest in head-mounted display technologies (e.g., smart glasses) has created new opportunities for egocentric sensing. Although such devices are still in the early stages of adoption, their potential has spurred substantial research on egocentric vision indoors, enabling general-purpose egocentric assistants \cite{grauman_ego4d_2022, plizzari2024outlook, yang_egolife_2026}. Extending these capabilities to dynamic urban environments remains an important but underexplored challenge. Addressing this gap is central to proactive AI systems for pedestrian assistance, including navigation support \cite{wang_egonav_2024, qiu_egocognav_2025, pan2025lookout} and specialized aids for visually impaired users \cite{haghighi2024heads}. 

\begin{figure}[t]
    \centering
    \includegraphics[width=\linewidth]{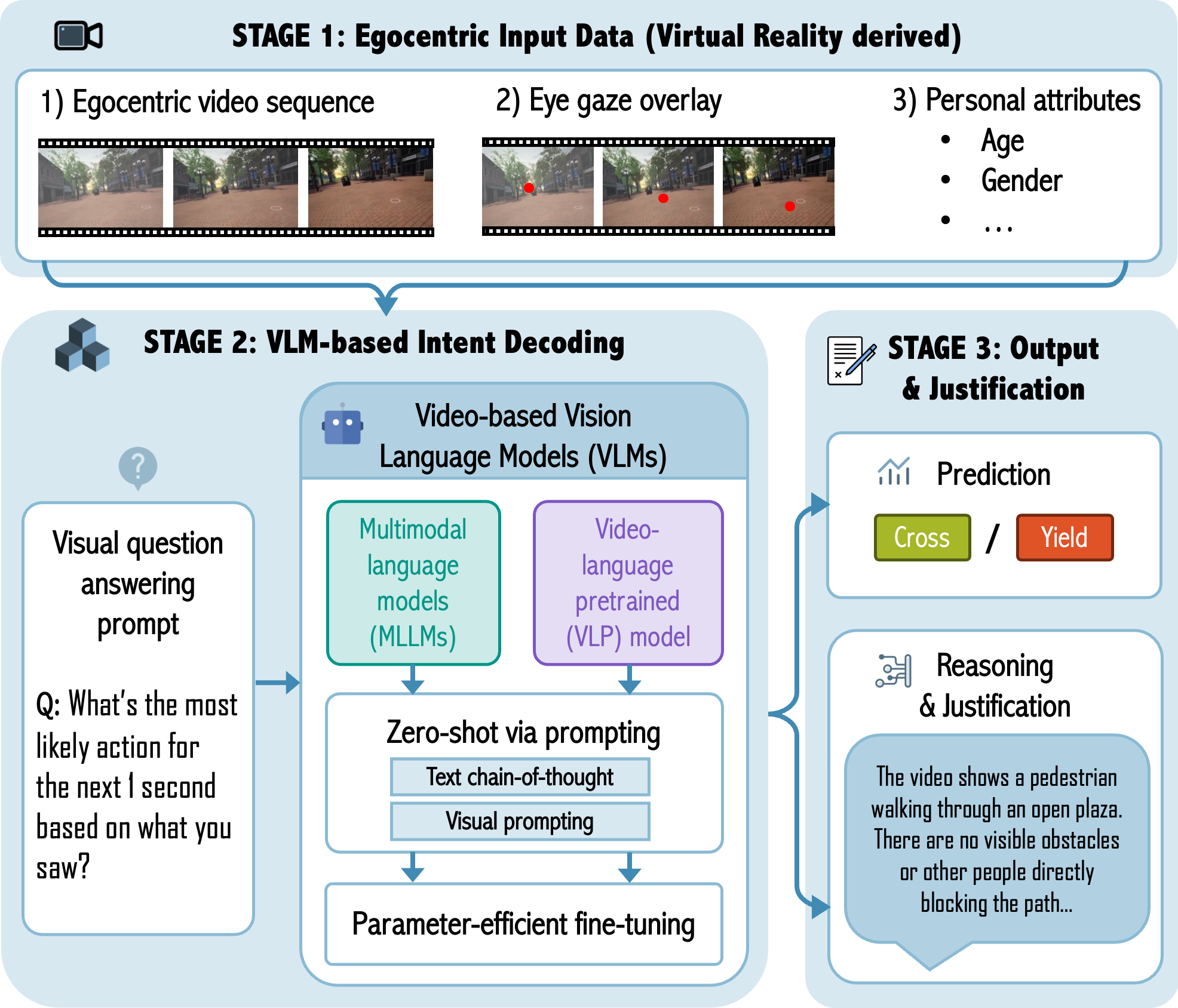}
    \caption{A pipeline of our approach.}
    \label{fig:overview}
\end{figure}

Existing research on egocentric prediction has largely focused on motion forecasting. Prior work has studied egocentric trajectory prediction in crowded spaces \cite{qiu_egocentric_2022} and on sidewalks \cite{singh_krishnacam_2016, pan2025lookout}, typically using task-specific architectures such as convolutional neural networks \cite{singh_krishnacam_2016}, diffusion models \cite{wang_egonav_2024}, and transformer-based approaches \cite{qiu_egocognav_2025}. These methods often use heterogeneous inputs, ranging from monocular views \cite{pan2025lookout, qiu_egocognav_2025} and stereo images \cite{park2016egocentric} to panoramic videos \cite{wang_egonav_2024}. Such specialization can limit transferability and generalization to diverse, real-world urban scenarios.

To address these limitations, we explore the potential of pretrained vision--language models (VLMs) for egocentric intention prediction by reformulating the problem as a visual question answering (VQA) task.
We ask: \textbf{Can we infer pedestrians' future crossing intention from monocular first-person videos in urban environments?}
Specifically, we first evaluate the zero-shot reasoning capabilities of state-of-the-art VLMs to assess whether they already possess the perceptual and reasoning skills required for urban navigation. We then study the impact of parameter-efficient fine-tuning for adapting these models to traffic-specific prediction. Finally, by incorporating gaze signals and personal attributes, we investigate whether additional contextual cues can improve egocentric traffic reasoning beyond video-only perception.
An overview of our framework is presented in Fig.~\ref{fig:overview}.

The remainder of the paper is organized as follows: Sec.~\ref{sec:related_work} reviews related literature. Sec.~\ref{sec:methodology} and \ref{sec:experimental_setup} describe the methodology and experimental setup. Sec.~\ref{sec:results_n_analysis} present and analyze the results. We conclude the paper in Sec.~\ref{sec:conclusion} with limitations and future directions.

\section{Related Work}
\label{sec:related_work}

\subsection{Pedestrian Intention Prediction}

Pedestrian intention prediction is a longstanding research topic in intelligent transportation and autonomous driving \cite{sharma2022pedestrian}. Most existing approaches operate under an \emph{exocentric} setting---from vehicle-mounted sensors, roadside cameras, or bird's-eye views---and infer what a pedestrian is about to do based on externally observable cues. 

In a typical pipeline, models combine multimodal inputs (e.g., tracked trajectories, body pose, and vehicle kinematics) with feature extractors and a downstream decoder/classifier. As summarized by recent surveys \cite{zhang_pedestrian_2023, landry_predicting_2025}, representative architectures span convolutional and recurrent models, graph-based interaction reasoning, and transformer-based sequence modeling.
More recently, VLMs have emerged as a promising paradigm for pedestrian-related traffic understanding. Several works probe multimodal large language models (MLLMs) in zero-shot settings to assess their traffic-scene understanding and reasoning \cite{huang_gpt-4v_2024, ham_omnipredict_2024, zambare2025seeing}. Other studies report gains via prompt engineering (e.g., hierarchical prompt templates \cite{azarmi_pedestrian_2025}), by using pretrained multimodal encoders as feature extractors \cite{uziel_optimizing_2025, munir_pedestrian_2025}, or by leveraging proprietary models for data annotation and teacher--student distillation \cite{gao_application_2025, ling2026vlmped}.

% Despite these advances, prior pedestrian intention prediction remains predominantly exocentric and models pedestrians as observed agents. This leaves an important gap: inferring intention from an \emph{egocentric} viewpoint that reflects the pedestrian's own perceptual evidence and decision making.

\subsection{Egocentric vision for behavior understanding}

% Egocentric vision studies videos captured from wearable sensors and cameras. 
The emergence of large-scale egocentric datasets has catalyzed research on general-purpose \emph{egocentric video understanding}. Many widely used datasets predominantly capture indoor daily activities, such as EPIC-KITCHENS \cite{damen_scaling_2018} featuring kitchen activities, Charades-Ego \cite{sigurdsson2018actor} consisting of everyday tasks, and Ego4D \cite{grauman_ego4d_2022} with more diverse first-person videos, still largely centered on household activities. These resources have enabled research problems including human--object interaction understanding, action recognition, and anticipatory/predictive modeling.
Video--language pretraining (VLP) approaches aim to learn aligned representations that better account for the characteristics of first-person videos (e.g., frequent viewpoint changes and strong ego motion). For example, EgoVLPv2 learns cross-modal representations from large-scale egocentric video--text data \cite{pramanick_egovlpv2_2023}, and recent efforts explore building foundation models tailored to egocentric video \cite{pei2024egovideo}. 
% In parallel, the community has begun to emphasize long-term egocentric understanding with memory-like capabilities (e.g., habit tracking and task support) \cite{yang_egolife_2026}.

% More recently, vision--language models (VLMs) and video--language pretraining have been extended to egocentric videos, enabling language-conditioned understanding beyond fixed label spaces. For example, EgoVLPv2 learns aligned representations from large-scale first-person video and text data \cite{pramanick_egovlpv2_2023}, and multimodal LLMs have demonstrated growing capability in open-ended egocentric reasoning \cite{patel_advancing_2025, peng2025eye}. Despite this progress, most egocentric VLM benchmarks and training data remain biased toward indoor activity understanding, whereas outdoor traffic interactions demand distinct reasoning about moving vehicles, right-of-way negotiation, and short-horizon action selection.

Despite rapid progress on indoor activity understanding, egocentric \emph{urban} understanding with highly dynamic, safety-critical interactions remains comparatively underexplored.
One relevant line of work is \emph{egocentric motion prediction} using deep learning, which forecasts the future movement of the camera wearer. Early approaches explored future localization from egocentric images or stereo inputs, for example via nearest-neighbor retrieval from a single egocentric image \cite{singh_krishnacam_2016} or stereo-based EgoRetinal representations \cite{park2016egocentric}. More recent methods predict wearer trajectories and head motion in crowded and urban environments \cite{qiu_egocentric_2022, pan2025lookout}, including diffusion-based modeling of future trajectories from panoramic videos \cite{wang_egonav_2024} and transformer-style sequence modeling with cognition-aware uncertainty \cite{qiu_egocognav_2025}. Wearable-camera datasets developed for assistive mobility further highlight the value of pedestrian-centric sensing for motion forecasting \cite{haghighi2024heads}. In our work, we move beyond motion forecasting and study short-horizon egocentric \emph{intention} decoding in traffic interactions using VLMs.

\subsection{Video question answering}

Video question answering (VideoQA) is a central task in video-language understanding, where models answer natural-language questions about dynamic visual content. Existing VideoQA tasks are commonly categorized into factoid and inference questions \cite{zhong2022video}. Factoid questions focus on directly observable evidence, such as locations, objects, attributes, actions, and activities, and mainly evaluate question understanding and visual recognition \cite{zhong2022video,xu2017video,jang2017tgif,yu2019activitynet}.

Recent benchmarks increasingly emphasize inference-oriented VideoQA, which requires reasoning over relations among objects, actions, events, states, and agents. These benchmarks evaluate temporal reasoning, such as action order, repetition, and state transitions \cite{jang2017tgif,xiao2021next}; spatial and compositional spatio-temporal reasoning over object-action relations and event structures \cite{jang2017tgif,grunde2021agqa}; causal reasoning through explanatory and predictive questions \cite{xiao2021next,li2022from}; and counterfactual reasoning through hypothetical ``what-if'' questions \cite{li2022from,jia2022egotaskqa}.

In egocentric settings, VideoQA further supports the evaluation of goal-directed task understanding, including action dependencies, action effects, intents, goals, and agents' beliefs \cite{jia2022egotaskqa}. Recent work also studies intent-oriented VideoQA in both third-person and egocentric videos, including context-aware intent reasoning and gaze-guided egocentric intent understanding \cite{li2023intentqa,peng2025eye}.

\section{Methodology}
\label{sec:methodology}

\subsection{Problem definition}

We study pedestrian crossing intention prediction from egocentric observations. Given a 2-second egocentric video clip $V$ and optional contextual information $C$, the goal is to predict whether the pedestrian will \emph{cross} or \emph{yield} within a 1-second future horizon. Formally, the task is defined as
\begin{equation}
    y = \mathcal{F}_{\Theta}(V, C),
\end{equation}
where $V \in \mathbb{R}^{T \times H \times W \times 3}$ denotes the input video with $T$ frames of spatial resolution $H \times W$, $C$ denotes additional contextual information described in Sec.~\ref{sec:contex}, and $y \in \mathcal{Y}=\{\text{cross}, \text{yield}\}$ denotes the predicted crossing intention.

We formulate intention prediction as a closed-ended visual question answering (VQA) task in order to leverage the visual understanding, language grounding, and reasoning priors of large pretrained VLMs. Specifically, the model receives the video $V$, additional contextual information $C$, and an intention query $Q$, and selects an answer from the predefined answer set $\mathcal{Y}$:
\begin{equation}
    A = \text{argmax}_{a \in \mathcal{Y}} p_{\Theta}(a \mid V, C, Q),
\end{equation}
where $A$ is the predicted answer and $p_{\Theta}$ denotes the answer probability estimated by the VLM.

\subsection{Model selection}
We consider two complementary families of video-based VLMs. The first family comprises \textbf{MLLMs} that directly process visual (video) inputs and are pretrained on large and diverse corpora, yielding broad world knowledge and general reasoning ability.
To balance performance and computational efficiency, we adopt Qwen3-VL-8B-Instruct and its lightweight variant Qwen3-VL-2B-Instruct \cite{qwen3technicalreport}, as well as Qwen2.5-VL-7B-Instruct, which has shown strong performance on egocentric understanding benchmarks \cite{peng2025eye}. We also evaluate InternVL3 \cite{zhu2025internvl3} using InternVL3-2B and InternVL3-8B. For the 7B and 8B models, we consider both 16-bit and 8-bit quantization.

The second family comprises \textbf{VLPs}, a de facto paradigm for video - text tasks. These models explicitly pretrain cross-modal representations on egocentric video and question answering data to improve video--text alignment \cite{pramanick_egovlpv2_2023}. However, their training data is often dominated by indoor activities \cite{damen_scaling_2018, grauman_ego4d_2022}. We select GroundVQA \cite{di_grounded_2024} (hereafter \textit{VLP}), a state-of-the-art architecture with dual video and language encoders and a cross-modal fusion module.

\subsection{Prompt design for VLMs}
We evaluate three prompting strategies, namely, standard prompting, text Chain-of-Thought (CoT) prompting \cite{wei_chain--thought_2022}, and visual prompting, because they probe complementary model behaviors that are all relevant to traffic-scene intention decoding. Standard prompting measures the model's out-of-the-box ability to map egocentric perception to a discrete action from minimal instructions, and therefore serves as our primary zero-shot reference. Text CoT prompting tests whether eliciting explicit intermediate reasoning in the language domain improves decision making beyond direct classification. Visual prompting, in contrast, injects task-relevant visual cues directly into the input to encourage spatial grounding and temporal consistency without modifying model parameters \cite{yang2023set,yang2023fine}.

For \textbf{standard prompt}, to enforce a structured response and fairly assess model capabilities, we design a constrained prompt: \prompt{``What is your most likely action in the next 1 second based on what you saw in the egocentric video for the past 2 seconds? Choose one option: (A) cross (B) yield.''}

For \textbf{text CoT prompting}, we consider a simple and an advanced variant.
First, we use \prompt{``Let's think step by step''} \cite{kojima_large_2023} to activate the chain-of-thought process. 
To further encourage multi-step reasoning, we use \prompt{``Analyze the egocentric video. First, describe the visual elements related to the crossing task. Second, evaluate attention presence, perceived proximity, and perceived risk. Third, explain the logic connecting these elements. Finally, provide the final answer. Output format: Reasoning: [maximum 5 sentences about your reasoning]. Answer: [just the letter and option].''}

\begin{figure*}[t]
\centering
    \begin{subfigure}{\textwidth}
    \centering
    \includegraphics[width=\textwidth]{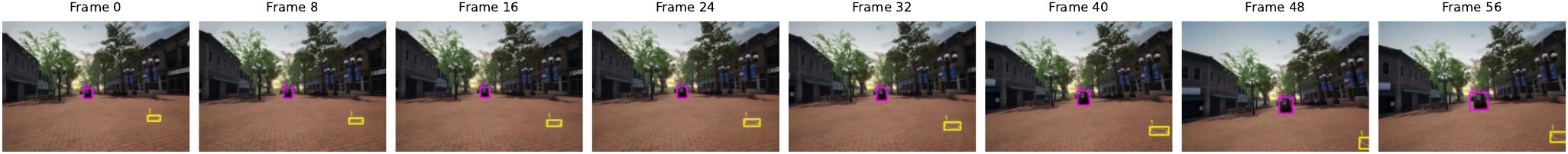}
    \caption{Example: Consistent tracking of the same objects over time.}
    \label{fig:stable_version}
    \end{subfigure}
    
    \begin{subfigure}{\textwidth}
    \centering
    \includegraphics[width=\textwidth]{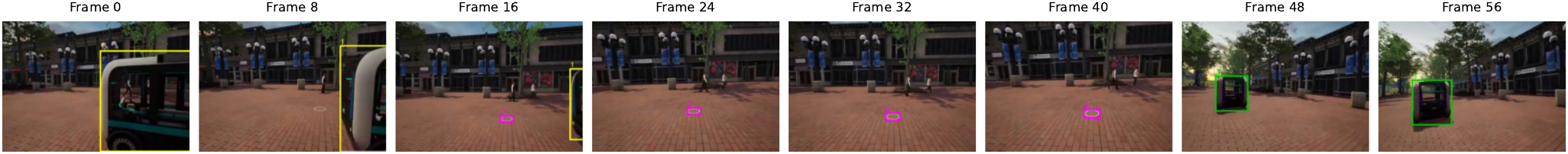}
    \caption{Example: Identification of new objects when turning head.}
    \label{fig:turning_version}
    \end{subfigure}

    \caption{Examples of visual prompt overlay on raw videos. }
    \label{fig:visual_prompt}
\end{figure*}

To construct the \textbf{visual prompt}, we augment the video input with offline visual cues, inspired by Set-of-Mark prompting for visual grounding \cite{yang2023set}. Specifically, we use the open-vocabulary detector GroundingDINO \cite{liu2024grounding} to detect task-relevant objects; in our setting, we focus on the automated vehicle and the white crossing-circle goal. The detections are refined via geometry-based filtering, cross-label suppression, and lightweight temporal tracking based on SORT \cite{bewley2016simple}, ensuring stable identifiers across sampled frames. We then render the processed detections as set-of-marks overlays, where tracked objects are annotated with persistent numeric IDs. Examples are shown in Fig.~\ref{fig:visual_prompt}. 

For video encoding, we consider two input representations. The first uses direct video inputs when the model natively supports video encoding. For the Qwen model family, the vision--language architecture employs multimodal rotary position embeddings to split videos into 3D chunks and to encode absolute temporal indices in the positional embedding \cite{qwen3technicalreport}. The second representation uses interleaved inputs, where each video is treated as an ordered sequence of individual frames and timestamps and image frames are alternated. 

\subsection{Representing contextual information}
\label{sec:contex}

We consider additional contextual information beyond the video input, including ego motion, vehicle motion \cite{li2025analyzing}, eye gaze \cite{li2026eye}, and personal attributes \cite{li2025analyzing}, all of which have been shown to be relevant to pedestrian intention prediction. \emph{Ego motion} describes the pedestrian's current position and speed, while \emph{vehicle motion} describes the vehicle's current position and speed. We also incorporate fine-grained eye-gaze signals to explicitly represent the pedestrian's line of sight \cite{li_challenges_2025}. Following \cite{peng2025eye, li_eye_2026}, we design two gaze-guided representations: (1) \emph{gaze direction}, where textual prompts describe the gaze orientation in the top-down view represented in degrees, indicating the direction to which the pedestrian is attending; and (2) \emph{gaze on screen}, where normalized gaze fixation coordinates on the image plane are represented as ratios in $[0,1]$. Personal attributes are incorporated by including demographic information in the prompt. For example: \prompt{``You are a 27-year-old female with a Master's degree or equivalent education level. Your dominant hand in daily life is your left hand. You walk every day...''.} In addition, we consider a visual gaze representation by rendering gaze fixations as red dots directly on the raw video frames. No overlay is rendered during saccades.

For time-varying contextual information, we further examine how it should be integrated into the prompt. Specifically, we consider two strategies. The first strategy directly appends the contextual information to the end of the standard prompt, under the assumption that the temporal trend can be inferred when the context is provided in a compact form; we refer to this strategy as \emph{preface}. The second strategy inserts contextual information alongside the corresponding timestamps when interleaved inputs are used. This design assumes that temporal alignment between contextual cues and video frames may help the model associate the context with specific visual evidence in the video; we refer to this strategy as \emph{interleaved}.

\subsection{Fine-tuning strategies}
We employ supervised fine-tuning (SFT) on annotated triplets of the form $(V, C, Q, A)$ to adapt pretrained models to egocentric traffic reasoning. 
To maintain computational efficiency while preserving pretrained knowledge, we adopt Low-Rank Adaptation (LoRA) \cite{hu_lora_2021}. LoRA freezes the original model parameters $\Theta_0$ and introduces trainable low-rank updates into selected transformer layers. For a weight matrix $W_0 \in \mathbb{R}^{d \times k}$, the adapted weight is given by
\begin{equation}
    W = W_0 + \Delta W, \qquad \Delta W = BA,
\end{equation}
where $B \in \mathbb{R}^{d \times r}$, $A \in \mathbb{R}^{r \times k}$, and $r \ll \min(d,k)$. During training, only the low-rank parameters $A$ and $B$ are updated.

The model is optimized using the standard cross-entropy loss over the closed answer set:
\begin{equation}
    \mathcal{L}
    = - \frac{1}{N} \sum_{i=1}^{N}
    \log p_{\Theta}(a_i \mid V_i, C_i, Q_i),
\end{equation}
where $a_i$ is the ground-truth answer for the $i$-th training example.

To analyze which components benefit most from domain adaptation, we selectively apply LoRA to different parts of the VLM: (1) the language encoders, (2) the cross-modal fusion modules, and (3) both the language and cross-modal components jointly. The vision encoder is kept frozen, following its role as a pretrained visual feature extractor in VLPs \cite{di_grounded_2024}.

\section{Experimental Setup}
\label{sec:experimental_setup}

Due to the limited availability of real-world egocentric data for pedestrian intention prediction in urban environments, we utilize a VR-based dataset from \cite{li_eye_2026}. This dataset captures egocentric pedestrian navigation alongside automated shuttles in a shared space. It provides synchronized egocentric videos, eye gaze tracking, demographic profiles, and movement trajectories.

To ensure data quality, we retain only critical interactions occurring prior to the intersection crossing point.
The data is segmented into 2-second observation windows and 1-second prediction horizons with a 0.5-second stride, yielding 6,047 QA samples.
To prevent data leakage and evaluate robust generalization, we partition the dataset at the participant-level into training, validation, and test sets using a 6:1:3 split.

\textbf{Labeling} 
We define the binary intention labels—``cross'' and ``yield''—based on pedestrian kinematic behavior within the 1-second future horizon.
Labels are determined by the duration for which the pedestrian speed exceeds a predefined crossing threshold \cite{li_eye_2026}.
Specifically, if the speed remains above the threshold for the majority of the horizon, the sample is labeled ``cross''; otherwise, it is labeled ``yield''. This results in 2,486 crossing and 3,561 yielding samples. To mitigate the resulting mild class imbalance, we apply random under-sampling to the majority class during subsequent training. 

\textbf{Baselines}
We establish a video-only deep learning baseline to forecast crossing intention.
Frame-level visual features are extracted using a pretrained CLIP backbone \cite{radford_learning_2021}, and the resulting feature sequence is processed by a transformer encoder followed by a classification head. This baseline is denoted as \textit{CLIP+Transformer} hereafter. 
% We validated the backbone and main architecture combination and found out the best baseline to achieve is a CLIP-based backbone with a LSTM-based architecture. 
The model is trained for up to 100 epochs using the Adam optimizer (learning rate 0.001, batch size 64) with early stopping to prevent overfitting. In addition, we consider two other simple baselines: (1) always predicting the majority class, and (2) random guessing. All baseline results are reported in Tab.~\ref{tab:zeroshot_egocentric}.

\textbf{Metrics}
We evaluate performance using accuracy, following standard practice in close-ended question answering (CloseQA) tasks \cite{patel_advancing_2025, peng2025eye}. To account for class imbalance, we additionally report the macro F1 score.
To mitigate choice-order bias, we randomize the associate between answer option and intention labels. Furthermore, we set the sampling temperature to 0 for deterministic generation, thereby isolating the model reasoning capabilities from stochastic variation.

\section{Results and Analyses}
\label{sec:results_n_analysis}

\subsection{Zero-shot performance using solely egocentric videos}

We first benchmark the zero-shot capability of pretrained MLLMs and a VLP model against standard baselines (Tab.~\ref{tab:zeroshot_egocentric}). With the standard prompt, none of the evaluated VLMs surpass the CLIP+Transformer baseline (Acc.=0.727, M-F1=0.724). Nonetheless, several MLLMs perform above random guessing, suggesting that they capture task-relevant semantics from egocentric traffic videos. In contrast, the VLP model exhibits a strong class bias, predicting ``yield'' almost exclusively, which yields majority-level accuracy (0.567) but low macro F1 (0.378).

Adding text CoT prompting to Qwen does not improve performance, while substantially increasing latency (from 0.1--0.3~seconds to roughly 2--4~seconds per sample; Tab.~\ref{tab:zeroshot_egocentric}). Moreover, the more elaborate multi-step CoT prompt increases the tendency to default to ``yield''. This deterioration observation aligns with recent evidence that explicit CoT may degrade visual-spatial reasoning in multimodal models \cite{dai2026latentomni}. A plausible explanation is that text CoT prompting amplifies linguistic priors while weakening reliance on fine-grained visual evidence.

Motivated by this observation, we further examine visual prompting via set-of-marks overlays. However, this strategy does not yield a measurable improvement over standard prompting. This may be attributed to: (i) imperfect detection and tracking that introduce inconsistent identities, (ii) limited visual diversity in the VR environment, and (iii) a suboptimal choice of prompted object categories for intention decoding.

\begin{table}[t]
\centering
\caption{Baselines, zero-shot VLMs, and fine-tuned models using only egocentric video. \textbf{Bold} indicates the best result within each group; \dag~denotes 8-bit quantization. ``Time'' reports per-sample inference latency in seconds.}
\label{tab:zeroshot_egocentric}
\resizebox{\linewidth}{!}{%
\begin{tabular}{lll!{\vrule}ccc}
\toprule
\textbf{Group} & \textbf{Model} & \textbf{Prompt} & \textbf{Acc.} & \textbf{M-F1} & \textbf{Time (s)} \\
\cmidrule(r){1-3} \cmidrule(l){4-6}
Baselines & Majority & --- & 0.567 & 0.362 & --- \\
          & Random   & --- & 0.500 & 0.497 & --- \\ 
          & CLIP+Transformer & --- & \textbf{0.727} & \textbf{0.724} & --- \\
\midrule
Zero-shot   & Qwen3-VL-2B & Standard     & 0.593 & 0.580 & 0.12 \\
            & Qwen3-VL-8B\dag & Standard & 0.579 & 0.559 & 0.32 \\ 
            & Qwen2.5-VL-7B\dag & Standard & \textbf{0.632} & \textbf{0.579} & 0.33 \\
            & Intern3VL-2B & Standard & 0.586 & 0.446 & 0.72 \\ 
            & Intern3VL-8B\dag & Standard & 0.594 & 0.475 & 1.35 \\ 
            & VLP  & Standard     & 0.567 & 0.378 & 0.01 \\
            
            \cmidrule(r){2-3} \cmidrule(l){4-6}
            & \qwensevenb\dag & CoT (simple) & 0.568 & 0.543 & 2.32 \\
            & \qwensevenb\dag & CoT (multi)  & 0.594 & 0.371 & 3.78 \\ 

            \cmidrule(r){2-3} \cmidrule(l){4-6}
            & \qwensevenb\dag & Visual prompt & 0.605 & 0.526 & 0.33 \\ 
            
\midrule
Fine-tuned & \qwentwob & Standard    & 0.777 & 0.773 & 0.12 \\ 
           & VLP  & Standard    & \textbf{0.789} & \textbf{0.787} & 0.007 \\
\bottomrule
\end{tabular}}
\end{table}

\begin{figure*}[t]
\centering
    \begin{subfigure}{\textwidth}
    \centering
    \includegraphics[width=\textwidth]{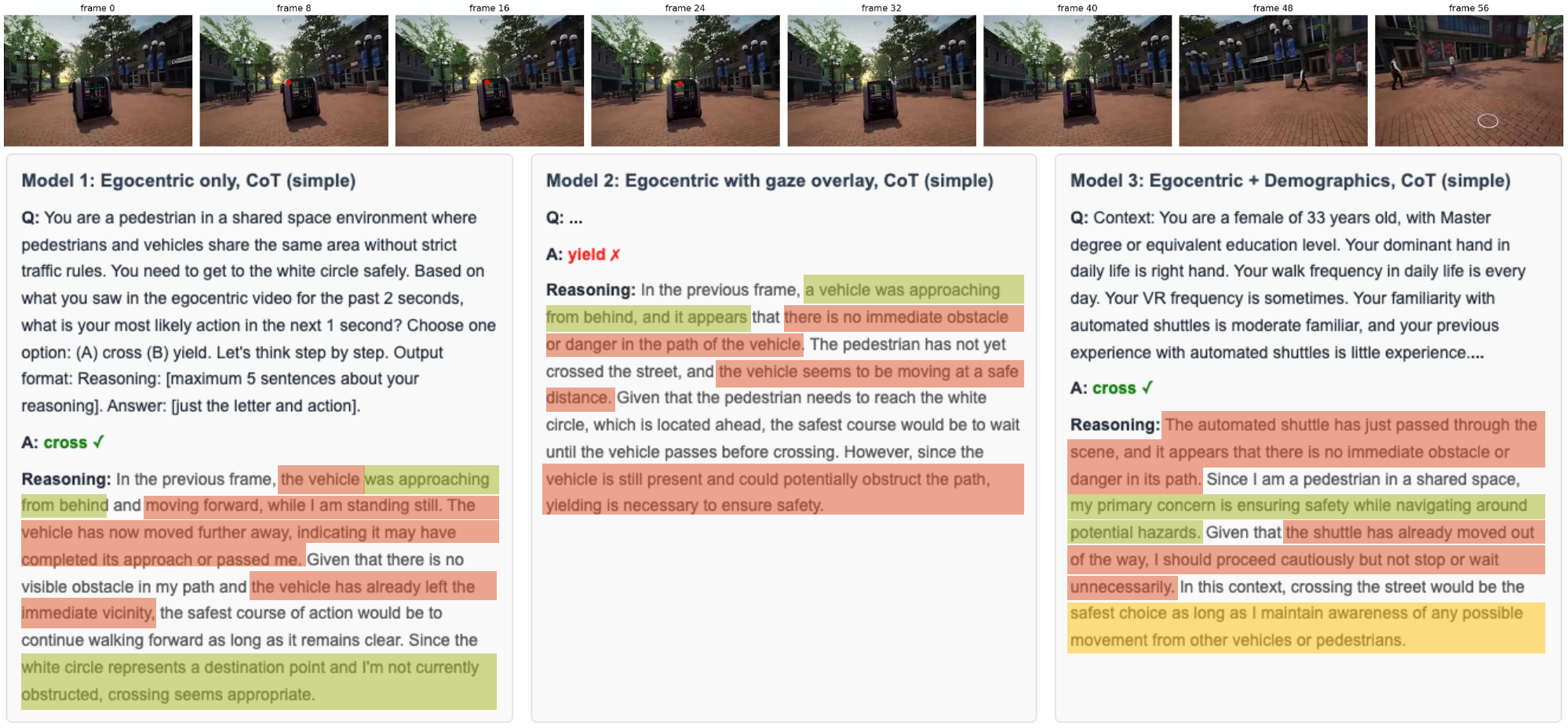}
    \caption{Failure case: Misinterpretation of vehicle dynamics. Both models fail to recognize the automated shuttle has stopped for the pedestrian. Without a clear understanding of the vehicle’s intent, the models incorrectly estimate the safety of the path.}
    \label{fig:bad_ex}
    \end{subfigure}
    
    \begin{subfigure}{\textwidth}
    \centering
    \includegraphics[width=\textwidth]{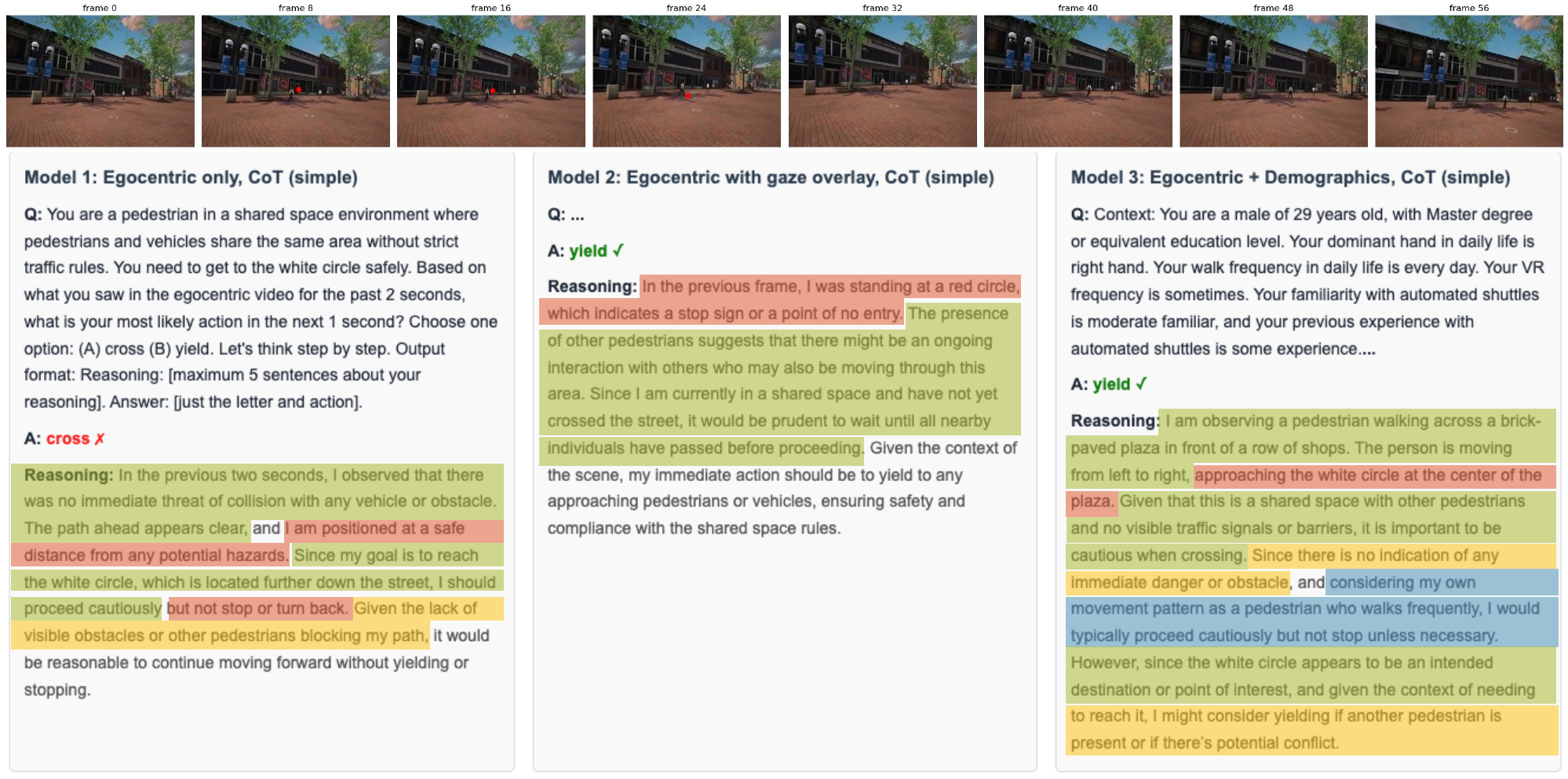}
    \caption{Success case: Gaze-informed hazard perception. The gaze-overlaid model correctly anticipates potential hazards. This shows gaze signals helps mimic human-like situational awareness. }
    \label{fig:good_ex}
    \end{subfigure}

    \caption{Qualitative analysis of Qwen outputs under CoT prompting. We contrast egocentric-only input (left) with gaze-augmented input (right). Highlight colors denote reasoning quality: red (incorrect), green (correct), yellow (partially correct), and blue (context-relevant).}
    \label{fig:reasoning}
\end{figure*}

We provide qualitative CoT examples in Fig.~\ref{fig:reasoning}. Although the model often identifies salient entities, it can still misread dynamic states. For example, in Fig.~\ref{fig:bad_ex}, the model recognizes the vehicle and the pedestrian's goal but incorrectly infers that the shuttle is moving and has already passed; in reality, the automated shuttle has stopped in front of the pedestrian.

\begin{figure}[t] % 't' suggests top placement
    \centering
    \begin{subfigure}{0.95\linewidth}
        \centering
        \includegraphics[width=\linewidth]{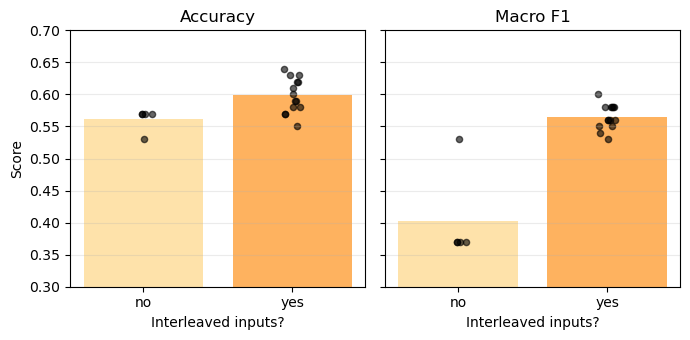}
        \caption{Effect of interleaved inputs.}
        \label{fig:ablation_zeroshot_interleaved_inputs}
    \end{subfigure}
    \hfill % Adds spacing between the two images
    \begin{subfigure}{0.95\linewidth}
        \centering
        \includegraphics[width=\linewidth]{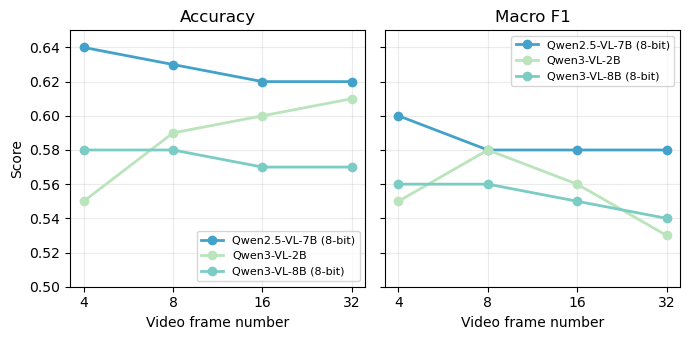}
        \caption{Effect of sampled video frame number.}
        \label{fig:ablation_zeroshot_video_frame_number}
    \end{subfigure}
    \caption{Zero-shot ablations on temporal input representation. (a) Effect of representing the clip as interleaved timestamps and frames. (b) Effect of the number of sampled video frames.}
    \label{fig:ablation_zeroshot}
\end{figure}

To further analyze zero-shot performance, we conduct an ablation study on the effects of video representation strategy and video frame sampling rate. The results are shown in Fig.~\ref{fig:ablation_zeroshot}.

\paragraph{Effect of video representation strategy} Fig.~\ref{fig:ablation_zeroshot_interleaved_inputs} shows that interleaving timestamps and frames generally outperforms non-interleaved inputs for the Qwen family. This resonates with recent findings that a simple timestamp--frame interleaving can improve temporal reasoning across models on localization-style tasks \cite{Meinardus_2025_ICCV}.

\paragraph{Effect of the sampled video frame numbers}

We assess how video sampling rate affects zero-shot performance. Fig.~\ref{fig:ablation_zeroshot_video_frame_number} shows two trends: for larger models (e.g., \qweneightb), increasing the number of frames from 4 to 32 generally degrades performance, whereas for the smaller model (\qwentwob), additional frames are beneficial. This suggests that larger models may be more sensitive to redundant or noisy inputs, while smaller models may benefit from additional visual evidence to compensate for limited capacity.

Overall, the zero-shot results indicate that models pretrained on broad web-scale data or predominantly indoor egocentric corpora can extract some decision-relevant cues in traffic-like interactions, but still fall short of strong task performance without adaptation.

\subsection{Fine-tuning performance using solely egocentric videos}

Building on the above findings, we apply parameter-efficient fine-tuning to adapt the pretrained VLMs to the target task while preserving their general knowledge. After adaptation, both models outperform the video-only CLIP+Transformer baseline (Tab.~\ref{tab:zeroshot_egocentric}): Qwen reaches Acc.=0.777, while VLP reaches Acc.=0.789. These improvements suggest that language-centric pretraining provides useful priors for mapping egocentric visual observations to discrete crossing decisions, but that domain adaptation remains necessary to capture traffic-specific interactions.

\begin{figure}[t] % 't' suggests top placement
    \centering
    \begin{subfigure}{0.48\linewidth}
        \centering
        \includegraphics[width=\linewidth]{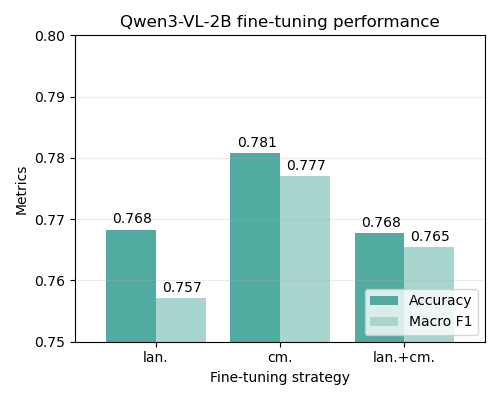}
        \caption{Qwen fine-tuning result.}
    \end{subfigure}
    \hfill % Adds spacing between the two images
    \begin{subfigure}{0.48\linewidth}
        \centering
        \includegraphics[width=\linewidth]{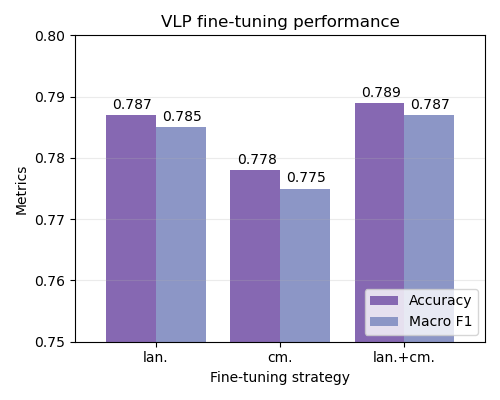}
        \caption{VLP fine-tuning result.}
    \end{subfigure}
    \caption{Comparison of parameter-efficient fine-tuning strategies for both VLMs. ``lan.'' tunes language modules, ``cm.'' tunes cross-modal modules, and ``lan.+cm.'' tunes both.}
    \label{fig:rq2_egocentric_only_ft}
\end{figure}

The optimal fine-tuning strategy varies across architectures (Fig.~\ref{fig:rq2_egocentric_only_ft}). For Qwen, updating only the cross-modal modules is sufficient to achieve strong performance, whereas VLP performs best when both the language and cross-modal components are adapted. Notably, VLP exhibits the larger improvement, suggesting greater adaptability to the target setting, potentially due to its egocentric pretraining.

% TODO: 
% \paragraph{Influence of rank}

\subsection{Impact of contextual information in zero-shot settings}

We next evaluate whether additional contextual cues improve zero-shot intention decoding. For the VLP model, adding contextual information leads to hallucinated outputs in approximately 5\% of cases; we therefore exclude it from this comparison. Tab.~\ref{tab:zeroshot_context_guided_summary} reports the results for \qwensevenb\ under different context configurations.

Three main observations emerge from the results. 
First, gaze-overlaid videos generally match or outperform raw videos across most context configurations, as indicated by the underlined entries in Tab.~\ref{tab:zeroshot_context_guided_summary}. Even without textual context, gaze overlay slightly improves accuracy over raw video input (0.640 vs. 0.632).
Second, adding a single contextual cue, such as gaze direction, gaze-on-screen coordinates, or vehicle motion, can degrade performance relative to the no-context setting. In contrast, combining these cues with ego motion yields more consistent gains. For example, with gaze-overlaid inputs, adding \emph{gaze on screen} together with \emph{ego motion} improves accuracy from 0.640 to 0.663. This suggests that ego-motion conditioning helps contextualize additional signals and stabilizes intention inference.
Third, \emph{ego motion} alone is already informative in the zero-shot setting, indicating that some contextual modalities may not be readily interpretable to the model without domain-specific training.

% \begin{table}[t]
% \centering
% \caption{Zero-shot performance with context guidance. For each setting, we report the best result over the context-FPS sweep. \textbf{Bold} number denotes the best performance in the group.}
% \label{tab:zeroshot_context_guided_summary}
% \begin{tabular}{ll|cc}
% \toprule
% Video Input & Context & Acc. & M-F1 \\
% \midrule
% Raw videos & - & \textbf{0.632} & \textbf{0.579} \\
%     & Gaze direction & 0.590 & 0.572 \\ 
%     & Gaze on screen & 0.592 & \textbf{0.579} \\
%     & Vehicle motion & 0.628 & 0.567 \\ 
%     \cmidrule(r){2-2} \cmidrule(l){3-4}
%     & Ego motion & 0.653 & \textbf{0.641} \\
%     & Ego motion, Gaze direction & \textbf{0.654} & 0.628 \\
%     & Ego motion, Gaze on screen & 0.646 & 0.636 \\
%     & Ego motion, Vehicle motion & 0.652 & 0.634 \\
% \midrule

% Gaze overlay & - & \textbf{0.640} & 0.574 \\
%     & Gaze direction & 0.622 & \textbf{0.592} \\ 
%     & Gaze on screen & 0.619 & 0.590 \\
%     & Vehicle motion & 0.636 & 0.564 \\ 
%     \cmidrule(r){2-2} \cmidrule(l){3-4}
%     & Ego motion & 0.640  & 0.588 \\
%     & Ego motion, Gaze direction & 0.657 & 0.614 \\
%     & Ego motion, Gaze on screen & \textbf{0.663} & \textbf{0.646} \\
%     & Ego motion, Vehicle motion & \textbf{0.663} & 0.637 \\
% \bottomrule
% \end{tabular}
% \end{table}

\begin{table}[t]
\centering
\caption{Zero-shot performance with contextual guidance. For each context configuration, we report the best result over the context sampling-rate sweep. \uline{Underline} marks the better video input (gaze overlay vs. raw) within the same context and metric; \textbf{bold} marks the best context overall for each metric.}
\label{tab:zeroshot_context_guided_summary}
\begin{tabular}{l|cc|cc}
\toprule
& \multicolumn{2}{c|}{Raw videos} & \multicolumn{2}{c}{Gaze overlay} \\
\cmidrule(l){2-3} \cmidrule(l){4-5}
Context & Acc. & M-F1 & Acc. & M-F1 \\
\midrule
-- & 0.632 & \uline{0.579} & \uline{0.640} & 0.574 \\
Gaze direction & 0.590 & 0.572 & \uline{0.622} & \uline{0.592} \\
Gaze on screen & 0.592 & 0.579 & \uline{0.619} & \uline{0.590} \\
Vehicle motion & 0.628 & \uline{0.567} & \uline{0.636} & 0.564 \\
\cmidrule(r){1-1} \cmidrule(l){2-3} \cmidrule(l){4-5}
Ego motion & 0.653 & 0.641 & \uline{\textbf{0.664}} & \uline{0.643} \\
Ego motion, Gaze direction & 0.654 & \uline{0.628} & \uline{0.657} & 0.614 \\
Ego motion, Gaze on screen & 0.646 & 0.636 & \uline{0.663} & \uline{\textbf{0.646}} \\
Ego motion, Vehicle motion & 0.652 & 0.634 & \uline{0.663} & \uline{0.637} \\
\bottomrule
\end{tabular}
\end{table}

Qualitative results (Fig.~\ref{fig:reasoning}) suggest that gaze overlay may enhance situational awareness, but in an implicit way. The generated reasoning rarely references gaze explicitly; nevertheless, as in Fig.~\ref{fig:good_ex}, the gaze-overlaid input can lead to more accurate projection of likely near-future events (e.g., inferring that an interaction with other agents may be ongoing, making it prudent to wait). Overall, these findings indicate that current pretrained models still struggle to reliably exploit context via prompting alone.

To further analyze these effects, we conduct an ablation study on context formatting and textual sampling rate when contextual cues are provided.

\paragraph{Effect of context formatting}
Tab.~\ref{tab:zeroshot_context_preface_interleaved} compares two formatting strategies: (i) appending contextual information as a preface and (ii) interleaving contextual information with video frames. The preface strategy consistently yields better results, whereas interleaving can substantially degrade performance. We speculate that interleaved text may disrupt the model's temporal visual processing, while a preface allows the model to condition globally on the contextual information before interpreting the frame sequence.

\begin{table}[t]
\centering
\caption{Zero-shot performance with contextual guidance (gaze-overlay video input), comparing two context formats: preface vs. interleaved. Both use identical inputs (same text sampling rate). \uline{Underline} marks the better format; \textbf{bold} denotes the best context overall for each metric.}
\label{tab:zeroshot_context_preface_interleaved}
\begin{tabular}{ll|cc}
\toprule
Context & Text format? & Acc. & M-F1 \\
\midrule
Ego motion & preface & \uline{\textbf{0.664}}  & \uline{\textbf{0.643}} \\
 & interleaved & 0.662  & 0.624 \\
\cmidrule(r){1-2} \cmidrule(l){3-4}
Ego motion, Gaze direction & preface & \uline{0.623} & \uline{0.538} \\
 & interleave & 0.618 & 0.532 \\
\cmidrule(r){1-2} \cmidrule(l){3-4}
Ego motion, Gaze on screen & preface & \uline{0.656} & \uline{0.612} \\
 & interleaved & 0.641 & 0.605 \\
\cmidrule(r){1-2} \cmidrule(l){3-4}
Ego motion, Vehicle motion & preface & \uline{0.663} & \uline{0.627} \\
 & interleaved & 0.623 & 0.544 \\
\bottomrule
\end{tabular}
\end{table}

\begin{figure}[t]
    \centering
    \includegraphics[width=\linewidth]{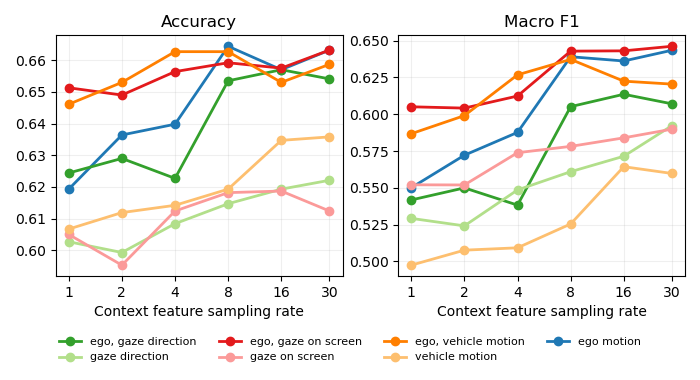}
    \caption{Effect of context sampling rate under zero-shot settings (gaze-overlay video input, prefaced context).}
    \label{fig:zeroshot_context_fps}
\end{figure}

\paragraph{Effect of context sampling rate} 
Because the frequency of interleaved context is tied to the video frame rate, we vary the textual sampling rate only for the preface strategy. The results are shown in Fig.~\ref{fig:zeroshot_context_fps}.
The accuracy curves suggest that the optimal sampling rate depends on the specific context type, although denser updates, roughly 4--30~Hz, generally perform better. For macro F1, performance tends to improve monotonically as text density increases. Across sampling rates, \emph{gaze on screen} and \emph{vehicle motion} are among the most beneficial cues and provide relatively consistent gains. One possible explanation is that ego and vehicle states can change rapidly, and higher-frequency updates provide more timely contextual conditioning.

\subsection{Impact of contextual information in fine-tuned settings}
\begin{figure}
    \centering
    \includegraphics[width=\linewidth]{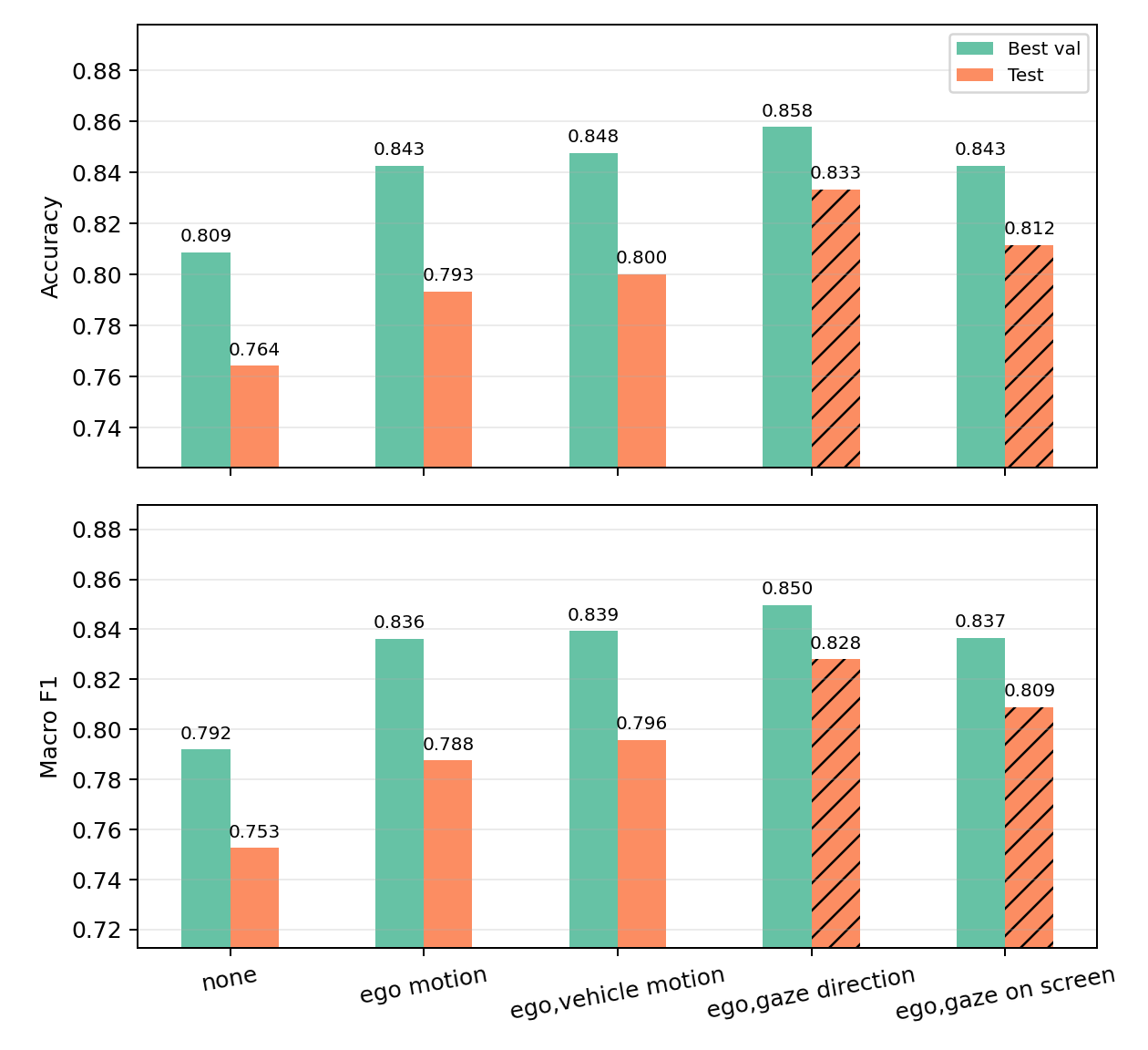}
    \caption{Validation vs. test performance of fine-tuned \qwentwob\ models under different context inputs.}
    \label{fig:context_val_test_comparison}
\end{figure}

To assess whether the model can learn to exploit additional contextual cues, we fine-tune \qwentwob\ using context-guided QA pairs (Fig.~\ref{fig:context_val_test_comparison}). Incorporating combined context improves performance compared with fine-tuning on egocentric video alone. On the validation set, different context combinations yield comparable performance, with \emph{ego motion + gaze direction} performing slightly better. On the test set, however, the differences become more pronounced. In particular, \emph{ego motion} alone, \emph{ego motion + gaze-on-screen}, and \emph{ego motion + vehicle motion} achieve similar validation accuracy, whereas \emph{ego motion + gaze-on-screen} produces a noticeably larger improvement on the test set. This suggests that gaze signals, when grounded by ego motion, can improve not only predictive accuracy but also generalization.

% \subsection{Discussion and Limitation}

% Our object detection and tracking algorithm might not be powerful enough to accurately and consistently label the same object, which might make the visual prompting underperform the standard prompting. 

\section{Conclusion}
\label{sec:conclusion}

This paper investigated the feasibility of using VLMs to decode pedestrian crossing intention from egocentric videos by formulating intention prediction as a closed-ended VQA task. Our experiments show that, although current off-the-shelf VLMs exhibit non-trivial zero-shot capabilities, they still fall short of a strong task-specific baseline and often struggle with higher-level traffic reasoning. In our setting, neither text-based chain-of-thought prompting nor set-of-marks visual prompting yields performance gains.
In contrast, parameter-efficient fine-tuning remarkably improves performance, with the best fine-tuned VLMs outperforming the CLIP+Transformer baseline by 9\% in accuracy.
Moreover, ego motion emerges as an informative contextual cue in zero-shot settings.
Additional contextual cues (vehicle motion and eye gaze) become more effective once the model is trained to exploit them: in particular, incorporating eye gaze improves both predictive accuracy and generalization.

Our study also has several limitations. First, the dataset is collected in a VR shared-space environment, which may not fully capture the long-tail complexity of real-world urban scenes. 
% (e.g., diverse lighting, sensor noise, unstructured behavior, and rare but safety-critical interactions)
Second, our video input is represented by uniformly sampled frames (often interleaved with timestamps), which introduces a key-frame selection problem and may miss brief but decisive cues. Third, qualitative failures suggest that understanding subtle vehicle intent (e.g., yielding vs. creeping) remains challenging.
Future work should therefore consider (1) traffic-specific multimodal pretraining at scale, (2) adaptive temporal sampling or event-driven frame selection, (3) richer and better-grounded visual representations, and (4) improved evaluation protocols on real-world egocentric data with more rare events and interpretable failure analysis.

\section*{Author contributions}

\noindent
\textbf{Danya Li}: Writing--original draft, Visualization, Validation, Methodology, Investigation, Formal analysis, Data curation, Conceptualization. 
% Wencan Mao: Methodology, Investigation, Formal analysis, Conceptualization. Francisco C. Pereira: Supervision, Resources, Conceptualization. Yu Xiao: Supervision, Resources, Conceptualization. 
\textbf{Xiang Su}: Writing--review \& editing, Supervision, Conceptualization. 
\textbf{Yan Feng}: Writing--review \& editing, Data curation. 
\textbf{Rico Krueger}: Writing--review \& editing, Supervision, Resources, Project administration, Funding acquisition, Conceptualization.

\bibliography{reference}

\end{document}